\documentclass[runningheads]{llncs}
\usepackage{graphicx}
\usepackage{tikz}
\usepackage{comment}
\usepackage{amsmath,amssymb} % define this before the line numbering.
\usepackage{color}
\usepackage{hyperref}
\usepackage{subfigure}
\usepackage{orcidlink}
\usepackage{booktabs}
\usepackage{multirow}
\usepackage{multicol}
\usepackage{makecell}
\usepackage{algorithm, algpseudocode}
\usepackage[algo2e,ruled,vlined,linesnumbered]{algorithm2e}
\usepackage[accsupp]{axessibility}  % Improves PDF readability for those with disabilities.

\begin{document}
% \linenumbers
\pagestyle{headings}
\mainmatter
\def\ECCVSubNumber{3419}  % Insert your submission number here

\title{EAGAN: Efficient Two-stage Evolutionary Architecture Search for GANs} % Replace with your title

\begin{comment}
\titlerunning{ECCV-22 submission ID \ECCVSubNumber} 
\authorrunning{ECCV-22 submission ID \ECCVSubNumber} 
\author{Anonymous ECCV submission}
\institute{Paper ID \ECCVSubNumber}
\end{comment}
%******************

% \begin{comment}
\titlerunning{EAGAN: Efficient Two-stage Evolutionary Architecture Search for GANs}
% \author{
% Guohao Ying$\dagger$\inst{2}\orcidID{0000-0001-9876-9253} \and
% Xin He$\dagger$\inst{1}\orcidID{0000-0001-8764-8157} \and
% Bin Gao\inst{3}\orcidID{0000-0001-5009-3514}\and
% Bo Han\inst{1}\orcidID{0000-0002-9226-0461} \and
% Xiaowen Chu$\ddagger$\inst{1,4}\orcidID{0000-0001-9745-4372}\thanks{$\dagger$: Equal contributions. $\ddagger$: Corresponding author (xwchu@ust.hk).}
% } 
%
% \author{
% Guohao Ying$\dagger$\inst{2}\orcidID{0000-0001-9876-9253} \and
% Xin He$\dagger$\inst{1}\orcidID{0000-0001-8764-8157} \and
% Bin Gao\inst{3}\orcidID{0000-0001-5009-3514}\and
% Bo Han\inst{1}\orcidID{0000-0002-9226-0461} \and
% Xiaowen Chu$\ddagger$\inst{4,1}\orcidID{0000-0001-9745-4372}\thanks{$\dagger$: Equal contributions. $\ddagger$: Corresponding author (xwchu@ust.hk).}
% } 

\author{
Guohao Ying$\dagger$\inst{2}\orcidlink{0000-0001-9876-9253} \and
Xin He$\dagger$\inst{1}\orcidlink{0000-0001-8764-8157} \and
Bin Gao\inst{3}\orcidlink{0000-0001-5009-3514}\and
Bo Han\inst{1}\orcidlink{0000-0002-9226-0461} \and
Xiaowen Chu$\ddagger$\inst{4,1,5}\orcidlink{0000-0001-9745-4372}\thanks{$\dagger$: Equal contributions. $\ddagger$: Corresponding author (xwchu@ust.hk).}
} 
% \author{
% Guohao Ying$\dagger$\inst{2} \and
% Xin He$\dagger$\inst{1}\thanks{$\dagger$: Equal contributions.} \and
% Bin Gao\inst{3}\and
% Bo Han\inst{1} \and
% Xiaowen Chu\inst{1,4}
% } 

\authorrunning{Guohao Ying, Xin He, Bin Gao, Bo Han, and Xiaowen Chu.}

\institute{
Hong Kong Baptist University, Hong Kong SAR, China \and
University of Southern California, USA \and
National University of Singapore, Singapore \and
The Hong Kong University of Science and Technology (Guangzhou), China \and
The Hong Kong University of Science and Technology, Hong Kong SAR, China
% \\ \email{xwchu@ust.hk}
}
% \end{comment}
%******************
\maketitle

\begin{abstract}

Generative adversarial networks (GANs) have proven successful in image generation tasks. However, GAN training is inherently unstable. Although many works try to stabilize it by manually modifying GAN architecture, it requires much expertise. Neural architecture search (NAS) has become an attractive solution to search GANs automatically. The early NAS-GANs search only generators to reduce search complexity but lead to a sub-optimal GAN. Some recent works try to search both generator (G) and discriminator (D), but they suffer from the instability of GAN training. To alleviate the instability, we propose an efficient two-stage evolutionary algorithm-based NAS framework to search GANs, namely \textbf{EAGAN}. We decouple the search of G and D into two stages, where stage-1 searches G with a fixed D and adopts the many-to-one training strategy, and stage-2 searches D with the optimal G found in stage-1 and adopts the one-to-one training and weight-resetting strategies to enhance the stability of GAN training. Both stages use the non-dominated sorting method to produce Pareto-front architectures under multiple objectives (e.g., model size, Inception Score (IS), and Fréchet Inception Distance (FID)). EAGAN is applied to the unconditional image generation task and can efficiently finish the search on the CIFAR-10 dataset in 1.2 GPU days. Our searched GANs achieve competitive results (IS=8.81$\pm$0.10, FID=9.91) on the CIFAR-10 dataset and surpass prior NAS-GANs on the STL-10 dataset (IS=10.44$\pm$0.087, FID=22.18). Source code: \href{https://github.com/marsggbo/EAGAN}{https://github.com/marsggbo/EAGAN}.

\end{abstract}

\section{Introduction}

Generative adversarial networks (GANs)~\cite{goodfellow2014generative} have obtained remarkable achievements on image generation tasks. A GAN consists of two networks (i.e., generator (G) and discriminator (D)) that contest with each other in a zero-sum game. G learns to generate semantic images from real data distributions, while D distinguishes real data from generated data. Since G and D have conflicting optimization objectives, GAN training is unstable and prone to collapse. Therefore, many efforts have been made to manually enhance architectures of GANs~\cite{dcgan,biggan}, but this requires much professional knowledge. Recently, neural architecture search (NAS) has proven to be effective in automatically finding superior models in various tasks~\cite{nas_survey,automl_he}, including GANs. The early NAS-GAN works \cite{autoGAN,agan} search only generator with a fixed discriminator to reduce search difficulty, but this may lead to a sub-optimal GAN. Although some recent works have searched both G and D, they suffer from the instability of GAN training. For example, AdversarialNAS~\cite{Adversarialnas}, which is the first gradient-based NAS-GAN, proposes an adversarial loss function to search G and D simultaneously, but the architectures of G and D are deeply coupled, which increases search complexity and the instability of GAN training. A subsequent gradient-based NAS-GAN work~\cite{AlphaGAN} also demonstrates that simultaneously searching both G and D hampers the search of optimal GANs. DGGAN~\cite{dggan} alleviates instability by progressively growing G and D but takes 580 GPU days to search on the CIFAR-10 dataset~\cite{cifar10}.

In this paper, we propose an efficient two-stage \textbf{E}volutionary \textbf{A}rchitecture search framework for \textbf{G}enerative \textbf{A}dversarial \textbf{N}etworks (\textbf{EAGAN}) on the unconditional image generation task. First, to alleviate the instability of GAN training during the search, we decouple the search of G and D into two stages. In stage-1, we fix the architecture of discriminator and search only generators. All generators are paired with the same discriminator, i.e., the candidate generators and the fixed discriminator are in a \textit{many-to-one} relationship. In stage-2, the best generator of stage-1 is used to provide supervision signals for searching discriminators. Specifically, in stage-2, we create multiple copies of the best generator architecture of stage-1, and each generator copy is paired with a different discriminator and trained independently. Thus, the generators and candidate discriminators of stage-2 are in a \textit{one-to-one} relationship. Because we indirectly evaluate the discriminators of stage-2 via IS (Inception Score \cite{IS}) and FID (Fréchet Inception Distance \cite{FID}) based on generators, the one-to-one strategy has a potential problem, i.e., if some generators have mode collapse at some time, then subsequently searched discriminators paired with these generators will be evaluated unfairly. To solve this problem, we propose the \textit{weight-resetting} strategy, where all generators inherit the weights of the best generator of the previous search round before a new search round starts. The results in Sec.~\ref{sec:ablation} show that our simple yet effective weight-resetting strategy can stabilize GAN searching. We summarize our contributions as follows.

\begin{enumerate}
    \item We greatly reduce the instability of GAN training by decoupling the search of generator and discriminator into two stages, where stage-1 and stage-2 adopt the \textit{many-to-one} and \textit{one-to-one} training strategy, respectively.
    \item We propose the \textit{weight-resetting} strategy, which is simple yet effective to avoid mode collapse when searching discriminators in stage-2 and ensure fair evaluations of different discriminators.
    \item EAGAN is efficient and takes 1.2 GPU days on the CIFAR-10 dataset to finish searching GANs. EAGAN achieves competitive results on the CIFAR-10 dataset and outperforms the prior NAS-GANs on the STL-10 dataset~\cite{stl10}.
\end{enumerate}

% \begin{figure}
%     \centering
%     \includegraphics[width=0.8\textwidth]{c10_s10_result2.pdf}
%     \caption{IS versus FID on the CIFAR-10 (left) and STL-10 (right) datasets. The green and red points indicate the results of searching only generators and searching both generators and discriminators, respectively. (Best viewed in color)}
%     \label{fig:visual_cifar10}
% \end{figure}

\section{Related Work}

\subsection{Generative Adversarial Network (GAN)}

% \begin{figure*}
%     \centering
%     \includegraphics[width=0.65\textwidth]{EAGAN3.pdf}
%     \caption{The pipeline of EAGAN. Different shapes and colors indicate different architectures and weights, respectively. In stage-1, a fixed discriminator guides search for the best generator $G^*$. In stage-2, each discriminator is paired with an independent generator having the same architecture as $G^*$. Before each round of stage-2, we weight-reset all generators the weights of the best generator of the previous round. (Best viewed in color)
%     % Then the best-performing generator in the stage-1 is replicated in $P$ copies. These copies are paired with different discriminators in the stage-2. For each round of stage-2, each generator weights are reset by the weights of the best generator of the previous round. Best viewed in color.
%     }
%     \label{fig:EAGAN}
% \end{figure*}

Generative Adversarial Networks (GANs) are first proposed in~\cite{goodfellow2014generative} and have been widely used in the various generation and synthesis tasks. A GAN comprises a generator (G) that generates plausible new data and a discriminator (D) that distinguishes the generator's fake data from real data. Suppose D and G are parameterized by $\theta$ and $\phi$, respectively, their loss functions are defined as

\begin{align}
L^D(\phi,\theta) &= -E_{x \sim p_{data}(x)}[\log D_{\theta}(x)] - E_{z \sim p(z)} [\log (1-D_{\theta}(G_{\phi}(z)))] \label{eq:D_loss}\\
L^G(\phi,\theta) &= E_{z \sim p(z)} [\log (1-D_{\theta}(G_{\phi}(z)))] \label{eq:G_loss}
\end{align}

\noindent where $p_{data}$ is the real data distribution and $p_z$ is a prior distribution. In other words, G and D play a min-max game with value function $V$, formulated below

\begin{align}
\min _{G} \max _{D} V(G, D)=& E_{x \sim p_{\text {data }}}[\log D(x)] +E_{z \sim p_{z}}[\log (1-D(G(z)))]
\label{eq:GAN}
\end{align}

\noindent The mix-max optimization incurs that GAN training suffers from multiple instability issues, such as mode collapse and gradient vanishing. To alleviate these problems, many efforts have been made~\cite{bissoto2019six} from the perspective of loss functions~\cite{wgan,improving_mmd_gan,bsgan}, normalization and constraint~\cite{wgan-gp,SNGAN}, conditional techniques~\cite{acgan,stylegan}, and validation methods~\cite{IS,FID}. Besides, architecture enhancements have been proven effective to improve GANs performance in many works~\cite{dcgan,biggan,progressiveGAN}.

% The commonly used metrics for GANs are Inception Score (IS)~\cite{IS} that measures the quality and diversity of the generated images, and Fréchet Inception Distance (FID)~\cite{FID} that captures the similarity between the generated images and real images.

\subsection{Neural Architecture Search (NAS)}

% Neural architecture search (NAS) that aims at automatic architecture design has achieved remarkable results in various fields~\cite{nas_survey,automl_he}, which can be formulated as a bilevel optimization problem below.

NAS aims at automatic architecture design and has achieved remarkable results in various fields~\cite{nas_survey,automl_he}. It can be formulated as a bilevel optimization problem as below

\begin{equation}
\begin{array}{ll} 
& \alpha^{*}=\arg \min _{\alpha} L_{\text {val }}\left(\alpha | w^{*}\right) \\
\text { s.t. } & w^{*}=\arg \min _{w} L_{\text {train }}(w | \alpha)
\end{array}
\end{equation}

\noindent where $L_{\text {train}}$ and $L_{\text {val}}$ indicate the training and validation loss; $w$ and $\alpha$ indicate the weight and architecture of neural network. This process aims to select the architecture $\alpha^*$ performing best on the validation set, conditioned on the optimal network weights $w$ on the training set. There are mainly four approaches in NAS: 1) Reinforcement learning (RL)~\cite{nas2016,enas} based methods train an RNN controller to generate neural networks; 2) Gradient-based methods~\cite{darts} apply softmax function to relax the discrete search space, allowing differential optimization of architectures; 3) Surrogate model-based optimization (SMBO)~\cite{pnas_liu18} builds a surrogate model of the objective function to predict the searched model's performance, which can substantially improve search efficiency; 4) Evolutionary algorithm (EA) based methods~\cite{amoebanet,cars} maintain and evolve a large population of neural architectures to produce the Pareto-front architectures.

% reinforcement learning (RL)~\cite{nas2016,nasnet_zoph17,enas}, gradient descent (GD)~\cite{darts,wu2019fbnet,Xu2020PC-DARTS}, surrogate model-based optimization (SMBO)~\cite{pnas_liu18}, and evolutionary algorithm (EA) based methods~\cite{amoebanet,cars,large_evolve,hiernas}.

\begin{table*}[!ht]
    \centering
    \scalebox{1}
    {
    \begin{tabular}{c|c|c|c|c}\hline
    Method & Type & search D? &Multi-objective? & Evaluation Metric(s)\\\hline
    
    AGAN~\cite{agan} & \multirow{3}{*}{RL} & × &×&IS\\
    AutoGAN~\cite{autoGAN} & & × & ×&IS\\
    E2GAN~\cite{offgan} & & × &$\surd$ & IS+FID$\dagger$\\\hline
    
    DEGAS~\cite{DEGAS} &  \multirow{3}{*}{Gradient} &×& ×&Loss\\
    AdversarialNAS~\cite{Adversarialnas} && $\surd$ & ×&Loss\\
    AlphaGAN~\cite{AlphaGAN} & & $\surd$& ×&Loss\\ \hline
    
    EGAN~\cite{EGAN} & \multirow{4}{*}{EA}& ×& $\surd$ & Loss \\
    EAS-GAN~\cite{EAS-GAN} & & x & x& Loss \\
    COEGAN~\cite{costa2019coevolution}& &$\surd$&×&FID (G); Loss (D)\\
    EAGAN & & $\surd$ &  $\surd$&Pareto-front(IS,FID,\#size)$\ddagger$\\\hline
    \end{tabular}
    }
    \caption{Comparison of our EAGAN and the existing NAS-GAN methods. The third column indicates whether the method supports searching discriminators. $\dagger$ indicates a linear combination of metrics. $\ddagger$ indicates the Pareto-front of multiple metrics.}
    \label{tab:NAS_GAN}
\end{table*}

\subsection{NAS for GANs}\label{sec:nas_gan}

Due to the great success of NAS in searching neural networks, many works have also applied NAS to search GANs, summarized in Table. \ref{tab:NAS_GAN}. AGAN~\cite{agan} and AutoGAN~\cite{autoGAN} are among the first RL-based NAS methods to search GANs, but they only use IS as the reward to guide the search. E2GAN~\cite{offgan} is rewarded by a linear combination of IS and FID. However, to avoid the notorious instability of GAN training, these early NAS-GAN methods only search generator (G) with a fixed discriminator (D) architecture, resulting in a sub-optimal GAN. AdversarialNAS~\cite{Adversarialnas} proposes to search G and D simultaneously in a differentiable way. However, it results in highly coupled architectures of G and D. The ablation study in~\cite{AlphaGAN} has demonstrated that simultaneously searching G and D would potentially increase the negative impact of inferior discriminators and hinder finding the optimal GANs. Liu et al.~\cite{dggan} propose to progressively grow the architectures of G and D in an alternating fashion, but this is only a remedy to alleviate the issue of architecture coupling and causes huge computational costs (580 GPU days on the CIFAR-10 \cite{cifar10} dataset).  COEGAN~\cite{costa2019coevolution} is very relevant to our work, which also uses an evolutionary algorithm to search G and D in two separate groups of architectures (called populations), but the two populations' architectures are coupled during the search. To reduce the search difficulty, COEGAN only explores a simple search space and experiments on a small dataset (MNIST \cite{mnist}). The final results show that COEGAN fails to outperform the previous human-designed GANs. In summary, since coupling G and D is not conducive to searching for the optimal GAN, we decouple them into two stages.

\section{Preliminary}

\subsection{Weight-sharing based Neural Architecture Search}

The early NAS methods first retrain the searched models from scratch and then evaluate their performance \cite{nas2016,amoebanet}, which obtains accurate evaluation but consumes huge resources, e.g., \cite{amoebanet} took 3,150 GPU days to search. To improve search efficiency, the weight-sharing strategy \cite{enas} was proposed to allow all subnets to share weights within a super network, so they can be evaluated without retraining by inheriting the weights from SuperNet. In our work, we also adopt the weight-sharing method to search generators and discriminators from SuperNet-G $\mathcal{N}_G$ and SuperNet-D $\mathcal{N}_D$, respectively. To simplify the notations, we use $\mathcal{N}$ to refer to both $\mathcal{N}_G$ and $\mathcal{N}_D$. Denote the loss of the $i$-th subnet $\mathcal{N}_i$ as $L_i$, and the weights of $\mathcal{N}$ as $W$. The gradients of SuperNet loss $L$ with respect to $W$ is

\begin{equation}
    \nabla_{W}L = \frac{1}{N}\sum_{i=1}^N\nabla_{W_i}L_i = \frac{1}{N}\sum_{i=1}^N\frac{\partial L_i}{\partial W_i}
\end{equation}

\noindent where $W_i$ is the weights of $\mathcal{N}_i$, and $N$ is the total number of subnets. However, it is not practical to accumulate all subnets' gradients in each batch. An alternative way is to use mini-batch subnets to update weights $W$. In our experiments, we find that randomly sampling one subnet (i.e., $N=1$) per batch can also work.

\subsection{Search Space}\label{sec:searchspace}

\begin{figure}
    \centering
    \includegraphics[width=\textwidth]{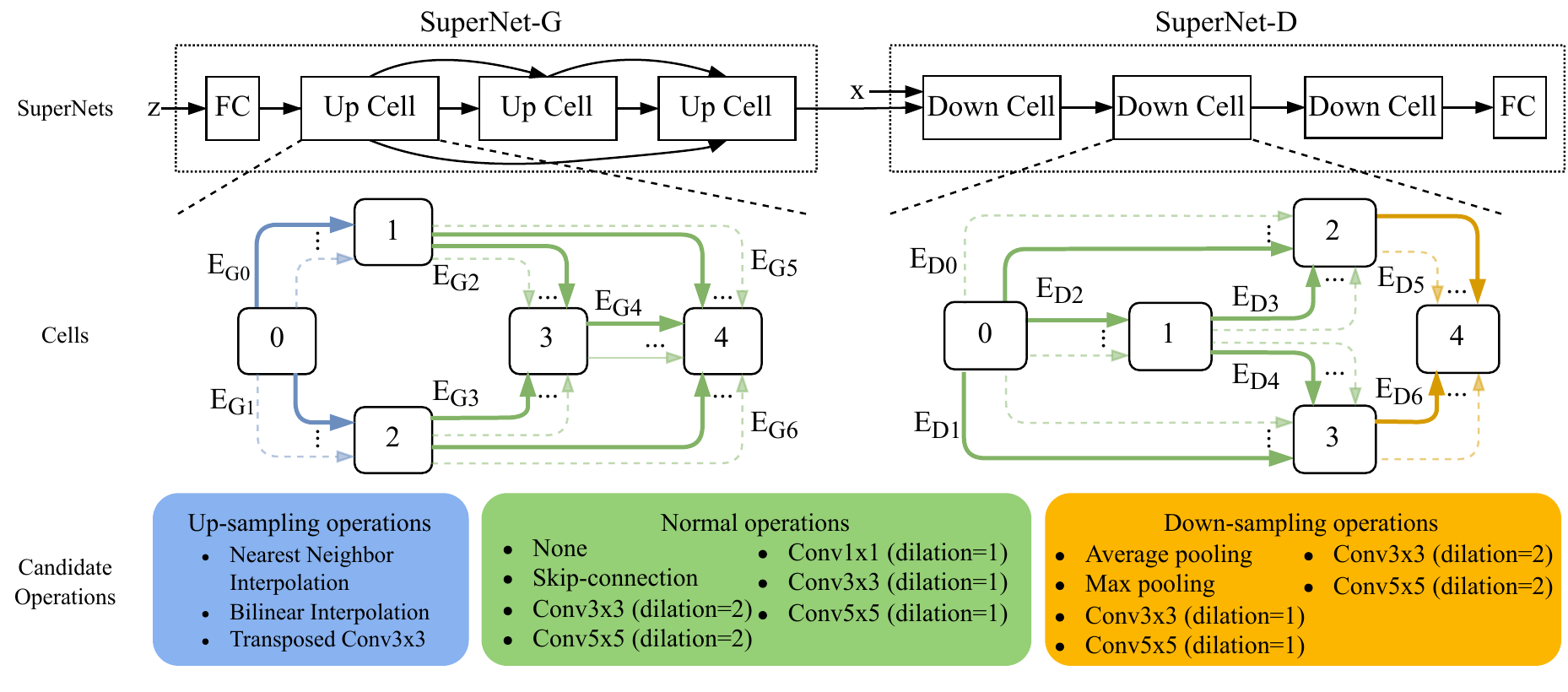}
    \caption{Overview of search space. $E_{G0}$ and $E_{G1}$ are up-sampling operations, $E_{D5}$ and $E_{D6}$ are down-sampling operations, and the other edges are normal operations.}
    \label{fig:searchspace}
\end{figure}

To ensure a fair comparison, we use the same search space as in~\cite{Adversarialnas} since it also searches both generators and discriminators. The search space is given in Fig.~\ref{fig:searchspace}.

% comprises two super networks (see Fig.~\ref{fig:searchspace}): SuperNet-G and SuperNet-D.

% In stage-1, all generators are searched from SuperNet-G. In stage-2, all discriminators are searched from SuperNet-D.

% During stage-1, all generators are sampled from SuperNet-G and SuperNet-D, respectively.  Similarly, all discriminators are sampled from SuperNet-D in stage-2.

% The search space of EAGAN comprises two super networks (see Fig.~\ref{fig:searchspace}): SuperNet-G $\mathcal{N}_G$ and SuperNet-D.

% As shown in Figure \ref{fig:EAGAN}, EAGAN searches generator ($G$) and discriminator ($D$) in two separate stages. Each stage involves architecture optimization and weights optimization. Inspired by~\cite{enas,cars}, we adopt the weight-sharing strategy to make the search efficient (Section \ref{sec:weightsharing}), which is carefully tailored for GANs. Specifically, generator and discriminator are searched from their own SuperNets: SuperNet-G $\mathcal{N}_G$ and SuperNet-D $\mathcal{N}_D$. The SuperNet contains all possible neural architectures. We adopt the search space as~\cite{Adversarialnas} for a fair comparison. The search space of EAGAN is shown in Fig. \ref{fig:searchspace}. We detail the structure of SuperNet-G and SuperNet-D as follows.

\textbf{SuperNet-G} $\mathcal{N}_G$ comprises a fully-connected (FC) layer and three Up-Cells. Each cell contains five ordered nodes (0-4), where node 0 is the output of the previous cell. There are multiple candidate operations between two nodes, each represented by an edge, and only one operation will be activated (solid edge). The edges $E_{G0}$ and $E_{G1}$ indicate up-sampling operations. The rest edges ($E_{G2}$ to $E_{G6}$) are normal operations, where ``None'' indicates no connection between two nodes. We encode each edge by a one-hot sequence. For example, [0,1,0] for edge $E_{G0}$ indicates that the bilinear interpolation operation is activated. \textbf{SuperNet-D} $\mathcal{N}_D$ comprises three Down-Cells and an FC layer. The Down-Cell is the inverted structure of the Up-Cell. The edges $E_{D0}$ to $E_{D4}$ are normal operations, and $E_{D5}$ and $E_{D6}$ are down-sampling operations. Thus, searching the architecture of G and D is transformed into searching a set of one-hot sequences.

\section{Methods}

% \subsection{Motivation}

% GAN training is notoriously unstable and prone to collapse; thus, the early NAS-GAN works only search generator (G). Although AdversarialNAS \cite{Adversarialnas} has made efforts to search G and D simultaneously in a differentiable way, the architectures of G and D are deeply coupled, which would potentially increase the negative effect of inferior discriminators and hinder the search process, as confirmed in a subsequent work \cite{AlphaGAN}. Liu et al.~\cite{dggan} propose to progressively grow the architectures of G and D, but suffers from huge computational costs (580 GPU days on the CIFAR-10 dataset). COEGAN~\cite{costa2019coevolution} is very relevant to our work, which also use evolutionary algorithm to search G and D. The difference is that . Therefore, it is still a challenge to search both G and D efficiently and effectively.

% In this section, we first describe the two-stage searching scheme and present details of the weight-resetting strategy. Then we introduce the search space of EAGAN and the weight sharing strategy. Finally, we illustrate the evolution operators.

% The pipeline of our two-stage EAGAN is given in Fig.~\ref{fig:EAGAN}. Both stages consist of two steps: weights training and evolutionary architecture search, where the latter step is the same so we introduce it together in Sec. \ref{sec:evolution}.

EAGAN comprises two stages, each having two steps: \textit{weights training} and \textit{architecture evolution}. The \textit{many-to-one} and \textit{one-to-one} training strategies tailored for two stages are detailed in Sec. \ref{sec:stage1} and Sec. \ref{sec:stage2}, respectively. Sec. \ref{sec:evolution} describes the steps for evolving architectures, which is the same in both stages.

% Sec. \ref{sec:searchspace} presents the search space.

% to alleviate the instability of GAN training as well as to improve the search efficiency

\subsection{Stage-1: Searching Generator}\label{sec:stage1}
% \subsubsection{Stage-1: Searching Generator}

% \subsubsection{Many-to-One GAN Training.} 

\begin{figure}
    \centering
    \includegraphics[width=\textwidth]{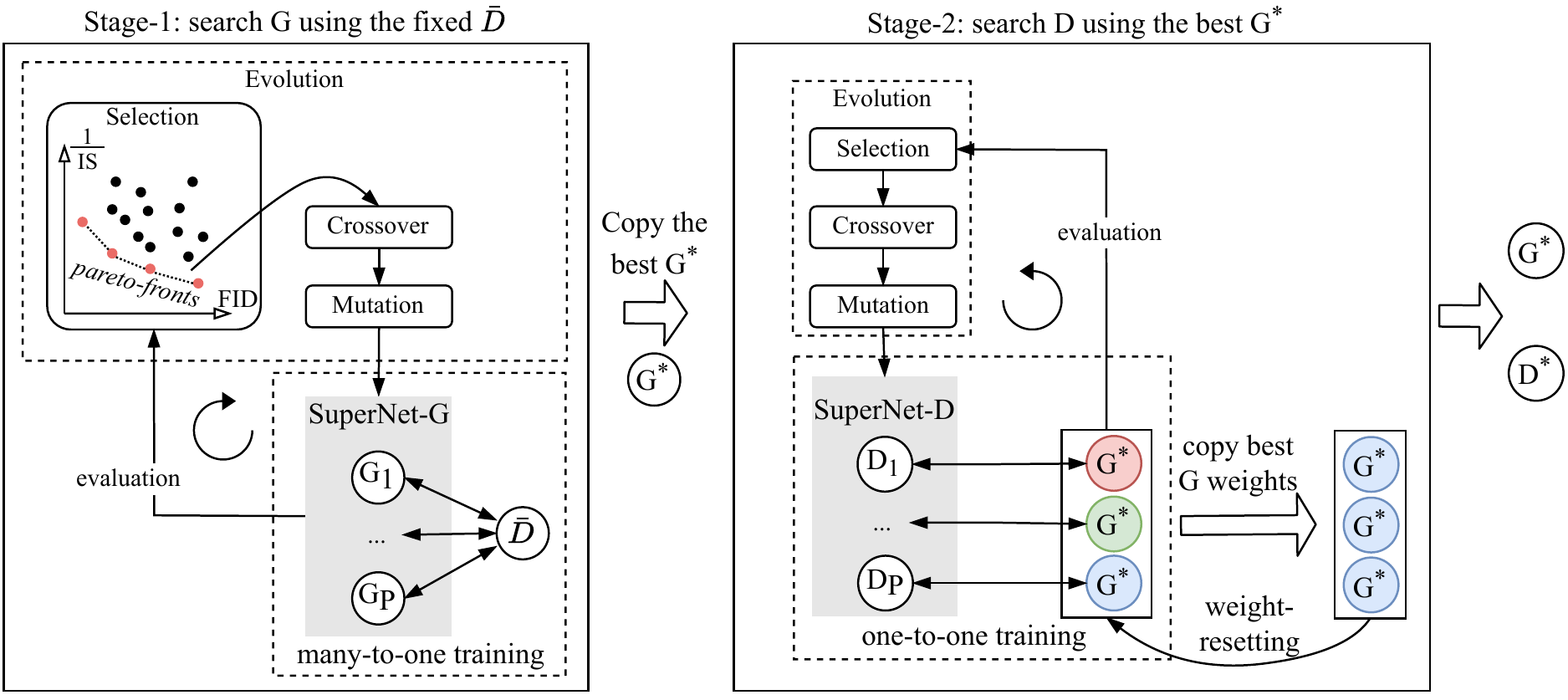}
    \caption{Two-stage pipeline of EAGAN. 
    % In stage-1, a fixed discriminator is used to guide the search of the best generator $G^*$. In stage-2, each discriminator is paired with an independent generator having the same architecture as $G^*$. Before starting a new  round of search in stage-2, we weight-reset all generators with the weights of the best generator of the previous round. Different colors of G indicate different weights.
    % Then the best-performing generator in the stage-1 is replicated in $P$ copies. These copies are paired with different discriminators in the stage-2. For each round of stage-2, each generator weights are reset by the weights of the best generator of the previous round. Best viewed in color.
    }
    \label{fig:EAGAN}
\end{figure}

\textbf{Many-to-One GAN Training.} As shown in Fig. \ref{fig:EAGAN} (left), in stage-1, we search generators (G) with a fixed discriminator (D) that has 0.91M parameters and the same architecture as that of~\cite{Adversarialnas}. We adopt the \textit{many(G)-to-one(D)} training strategy. Specifically, the fixed discriminator $\bar{D}$ is denoted by architecture and weights variables, i.e., $\bar{D} \sim (\bar{\beta},w_{\bar{D}})$. During each round, we produce $P$ candidate generators to form the \textit{population-G} $\mathcal{A}_G$, where all candidate generators share the weights $W_G$ of SuperNet-G, and each candidate $G_{ i }$ is parameterized with architecture and weights variables, i.e., $G_{ i } \sim (\alpha_{ i },w_{G_{ i }})$, where $w_{G_{ i }}=W_G(\alpha_{ i })$. We then pair each candidate generator with the fixed discriminator $\bar{D}$ to form $P$ GANs, i.e., $\{(G_1,\bar{D}),...,(G_P,\bar{D})\}$. Stage-1 can be formalized as below

% Stage-1 comprises multiple rounds, and each round comprises two steps: 1) many-to-one GAN training (Eq. (\ref{eq:stage1_eq2})$\sim$(\ref{eq:stage1_eq3})); and 2) evolving G architectures (Eq. (\ref{eq:stage1_eq1})), formalized below

% As shown in Fig. \ref{fig:EAGAN} (left), we search generators (G) with a fixed discriminator (D) in stage-1. D is denoted by architecture and weights variables, i.e., $\bar{D} \sim (\bar{\beta},w_{\bar{D}})$. All candidate generators share weights ($W_G$) of SuperNet-G. Similar to D, each candidate $G_{ i }$ is also parameterized with architecture and weights variables, i.e., $G_{ i } \sim (\alpha_{ i },w_{G_{ i }})$, where $w_{G_{ i }}=W_G(\alpha_{ i })$. Stage-1 comprises multiple rounds, and each round comprises two steps: 1) many-to-one GAN training (Eq. (\ref{eq:stage1_eq2})$\sim$(\ref{eq:stage1_eq3})); and 2) evolving G architectures (Eq. (\ref{eq:stage1_eq1})), formalized below

\begin{align}
\alpha^{*} &=\arg \min _{\alpha_{ i }} \{ V_{val}\left(\alpha_{ i } \mid w_{G_{ i }}^{*}, w_{\bar{D}}^{*}, \bar{\beta}\right) , i\in\{1,...,P\}\} \label{eq:stage1_eq1} \\
\text { s.t. } \quad w_{G_{ i }}^{*} &=\arg \min _{w_{G_{ i }}} \,  E_{z \sim p(z)}\left[\log \left(1-\bar{D}\left(G_{ i }(z)\right)\right)\right] \label{eq:stage1_eq2} \\
w_{\bar{D}}^{*} &=\arg \max _{w_{\bar{D}}} \sum_{i=1}^{P}  E_{x \sim p_{\text {data }}(x)}[\log \bar{D}(x)]+  E_{z \sim p(z)} [\log  (1-D (G_{ i }(z) ) ) ]  \label{eq:stage1_eq3}
\end{align}

% \begin{align}
% \alpha^{*} &=\arg \min _{\alpha_{ i }} \{ V_{val}\left(\alpha_{ i } \mid w_{G_{ i }}^{*}, w_{\bar{D}}^{*}, \bar{\beta}\right) , i\in\{1,...,P\}\} \label{eq:stage1_eq1} \\
% \text { s.t. } \quad w_{G_{ i }}^{*} &=\arg \min _{w_{G_{ i }}} E_{z \sim p(z)}\left[\log \left(1-\bar{D}\left(G_{ i }(z)\right)\right)\right] \label{eq:stage1_eq2} \\
% w_{\bar{D}}^{*} &=\arg \max _{w_{\bar{D}}} \sum_{i=1}^{P}  E_{x \sim p_{\text {data }}(x)}[\log \bar{D}(x)]+  E_{z \sim p(z)} [\log  (1-D (G_{ i }(z) ) ) ]  \label{eq:stage1_eq3}
% \end{align}

% \noindent In our work, Eq. (\ref{eq:stage1_eq1}) is implemented by the \textit{non-dominated sorting} method~\cite{nsga2} (detailed in Sec. \ref{sec:evolution}), which can find the optimal G architecture under multiple objectives (e.g., IS, FID, and model size) based on the validation set.

\noindent where the inner (Eq. (\ref{eq:stage1_eq2})$\sim$(\ref{eq:stage1_eq3})) is to optimize weights of $P$ GANs on the training set via the many-to-one strategy, and the outer (Eq. (\ref{eq:stage1_eq1})) is to obtain the optimal architecture of G according to the value function on the validation set (i.e., $V_{val}$). The inner and outer optimizations are solved by iterative procedures, outlined in Alg. \ref{alg:eagan}. These $P$ GANs share the same discriminator and are trained for multiple epochs for each round. To get a fair comparison between generators, for each training batch, we uniformly draw a generator from $P$ candidate generators and train it with the fixed discriminator (lines 4 to 10 in Alg. \ref{alg:eagan}). The many-to-one training mechanism can bring two benefits. First, the fixed discriminator $\bar{D}$ is trained with various generators, which can be viewed as an ensemble method to some extent, avoiding that $\bar{D}$ is over-fitted and much stronger than generators. Second, different generators are trained with the same discriminator, so we can fairly compare the performance of these generators to find the optimal one. Besides, a generator with mode collapse will not interfere with other generators because the selection step will eliminate it from the population (see Sec. \ref{sec:evolution}).

\subsection{Stage-2: Searching Discriminator}\label{sec:stage2}
% \subsubsection{Stage-2: Searching Discriminator with Weight-resetting Strategy}

% After stage-1, we obtain an optimal generator $ G^* $ with architecture $ \alpha^* $, conditioned on the fixed discriminator architecture $\bar{\beta}$.

After stage-1, we obtain an optimal generator $ G^* $ with architecture $ \alpha^* $. In stage-2, we use it to guide searching discriminators (D). There are two major challenges in searching D: the lack of evaluation metrics for discriminators and the instability of GAN training. Next, we describe our approaches to these two challenges.

% As shown in Fig. \ref{fig:EAGAN} (right), stage-2 also comprises multiple rounds of weights training and architecture evolution, but it has two main challenges, for which we propose corresponding solutions, as detailed below.

% However, we cannot directly use the same steps as stage-1 in stage-2, due to two main challenges described below.

% \subsubsection{One-to-One GAN Training.}

\textbf{One-to-One GAN Training.} Unlike generators, discriminators are difficult to be assessed directly. For example, the accuracy of discriminators does not reflect the overall performance of GANs, as high accuracy may indicate that generators are too weak to fool discriminators, and low accuracy may indicate that generator has mode collapse, with no way to analyze the real cause. Some works \cite{Adversarialnas,AlphaGAN,costa2019coevolution} use the reconstructed loss (e.g., Eq. (\ref{eq:D_loss})) to monitor discriminator, but the loss is not a reliable monitor metric as GAN training is a dynamic equilibrium process. An alternative solution is to \textit{indirectly} assess the discriminator via IS and FID metrics calculated based on a generator, so we cannot simply imitate the training strategy of stage-1 (e.g., many(D)-to-one(G)) in stage-2; otherwise, all discriminators are paired with the same generator and not comparable. To this end, we propose the \textit{one-to-one} training strategy. Specifically, we create $P$ copies of $ G^* $, each paired with a candidate discriminator from \textit{population-D} $\mathcal{A}_D$. Thus, we obtain $P$ GANs, i.e., $\{(G_i,D_i), i\in\{1,...,P\}\}$, where $G_i \sim (\alpha^*, w_{G_i})$ and $D_i \sim (\beta_i,w_{D_i})$. Each GAN is independently trained as a regular GAN via Eq. (\ref{eq:D_loss})$\sim$(\ref{eq:GAN}). Therefore, stage-2 can be formalized as follows

% During each search round, we produce $P$ candidate discriminators to form the \textit{population-D} $\mathcal{A}_D\}$. Unlike stage-1 where all candidate generators are paired with the same discriminator, in stage 2, we pair each candidate discriminator with independent generators having the same architecture to form P GANs, i.e., $\{(\alpha^*,\beta_{ 1 }),...,(\alpha^*,\beta_{ P })\}$. For example, G and D 
% Each pair of G and D is independently trained as a regular GAN (i.e., Eq. (\ref{eq:D_loss}) $\sim(\ref{eq:GAN})$).

\begin{align}
\beta^{*} &=\arg \min _{\beta_{ i }} \{ V_{val}\left(\beta_{ i } \mid w_{G_{ i }}^*, w_{D_{ i }}^{*}, \alpha^*\right) , i\in\{1,...,P\}\} \label{eq:stage2_eq1} \\
\text { s.t. } w_{G_{ i }}^*, w_{D_{ i }}^* &= \min _{G_{ i } } \max _{D_{ i }} E_{x \sim p_{\text {data }}(x)}[\log D_{ i }(x)] +E_{z \sim p(z)}[\log (1-D_{ i }(G_{ i }(z)))] \label{eq:stage2_eq2}
\end{align}

% \noindent Similar to stage-1, we uniformly train each pair of generator and discriminator via Eq. (\ref{eq:stage2_eq2}), ensuring each GAN trained equally (lines 21 to 27 in Alg. \ref{alg:eagan}).

% , the evaluation for discriminators we use IS and FID to indirectly assess the discriminator performance, and these two metrics are calculated based on the generator. In other words, if all discriminators are paired with the same generator, we cannot compare these discriminators. A possible solution is to pair each discriminator with an independent generator, which has the same architecture as $G^*$. However, a potential problem with this solution is that, with constant training and searching, the weights of each generator may be optimized in different directions, and some of them may have mode collapse. This will result in an unfair and biased evaluation for the discriminators paired with those collapsed generators. 

\begin{algorithm}[!ht]
    \SetAlgoLined
    \KwIn{SuperNet-G $\mathcal{N}_G$, SuperNet-D $\mathcal{N}_D$, population-G $\mathcal{A}_G$, population-D $\mathcal{A}_D$, population size $P=|\mathcal{A}_G|=|\mathcal{A}_D|$, multi-objective set $\mathcal{F}$, total search rounds $R$, each round contains $E$ epochs of training.
    }
    \KwOut{$G^*$ and $D^*$}
    
    $\bar{D}\sim(\bar{\beta},w_{\bar{D}}) \leftarrow$ Initialize a discriminator with weights $w_{\bar{D}}$ and fixed architecture $\bar{\beta}$;
    
    $\mathcal{A}_G^{(0)}=\{G_1^{(0)},...,G_P^{(0)}\}\leftarrow$ Warm-up($\mathcal{N}_G, \bar{D}$);
    
    $\{(G_i^{(0)},\bar{D}\}), i\in\{1,...,P\}\} \leftarrow$ Initialize $P$ GANs that share the same discriminator;

    \For{r=0:$R-1$}{
        \For{e=0:$E-1$}{
            \For{batch $x=\{x_1,...,x_m\}$ in training set}{
                Sample noise data $z=\{z_1,...,z_m\}$;
                
                Uniformly sample ${G_i^{(r)}}$ from $\mathcal{A}^{(r)}_G, i\in\{1,...,P\}$;
                
                Update weights of $\bar{D}$ via Eq. (\ref{eq:stage1_eq3});
                
                Update weights of ${G_i^{(r)}}$ via Eq. (\ref{eq:stage1_eq2});

            }
        }
        $\mathcal{A}^{(r)}_G\leftarrow$ Select Pareto-front generators under $\mathcal{F}$ based on validation set;
        
        % update $\mathcal{A}^{(r)}_G$ using NSGA-II based on $\mathcal{F}$
        
        $\mathcal{A}^{(r)}_G\leftarrow$  Crossover\&Mutation($\mathcal{A}^{(r)}_G$);
    }
    % $G^*=(\mathcal{N}^{(R)}_{G_k},W_{G_k}^{(R)}),k\in[1,P]\leftarrow$ the best generator
    
    $G^*\sim(\alpha^*, w_{G^*})\leftarrow$ the best generator with architecture $\alpha^*$ and weights $w_{G^*}$;
    % $G^*=(\mathcal{N}_{G^*},W_{G^*})\leftarrow$ the best generator
    
    % $W_{G_{best}}=W_{G^*}$
    
    % $G=\mathcal{N}^{(R)}_{G_k}$, which is the best among $\mathcal{A}^{(R)}_G,k\in[1,P]$
    
    % $W_{best}=W_G$
    
    $\mathcal{A}_D^{(0)}=\{{D_1^{(0)}},...,{D_P^{(0)}}\}\leftarrow{}$ Warm-up($G^*, \mathcal{N}_D$);
    
    % Initialize $P$ GANs $\{(G_{W_1},\mathcal{N}_{D_1}^{(0)}),...,(G_{W_P},\mathcal{N}_{D_P}^{(0)})\}$
    
    % Initialize $P$ generators $\{G_1,G_2,...,G_P\}$, where $G_i=(\mathcal{N}_{G^*},W_{G_i})$
    
    % $\{(\mathcal{N}_{G^*},\mathcal{N}_{D_1}^{(0)}),...,(\mathcal{N}_{G^*},\mathcal{N}_{D_P}^{(0)})\}$
    $\{(G_i,{D_i^{(0)}}), i\in\{1,...,P\}\} \leftarrow$ Initialize $P$ GANs, where $G_i$ is a copy of $G^*$;
    
    \For{r=0:$R-1$}{
        % Weight-resetting $W_{G_1}=...=W_{G_P}=W_{G^*}$
        % Weight-resetting $W_{G_1}=...=W_{G_P}=W_{G_{best}}$
        
        \For{e=0:$E-1$}{
            \For{batch $x=\{x_1,...,x_m\}$ in training set}{
                Sample noise data $z=\{z_1,...,z_m\}$;
                
                Uniformly sample a GAN $(G_i,{D_i^{(r)}})$ from $P$ GANs;
                
                % Calculate D loss $\sum^m_{j=1}H_D(x_j,z_j|G_i,\mathcal{N}^{(r)}_{D_i})$
                
                % ascending its stochastic 
                Update weights of $G_i $ and ${D_i^{(r )}}$ via Eq. (\ref{eq:stage2_eq2});
                
                % by gradient $\nabla\sum^m_{j=1}L_D(x_j,z_j|G_i,\mathcal{N}^{(r)}_{D_i})$
                
                % Sample noise data $z=\{z_1,...,z_m\}$ 
                
                % % Calculate G loss $\sum^m_{j=1}H_G(z_j|G_i\mathcal{N}^{(r)}_{D_i})$
                % %  descending its stochastic
                % Update weights of $G_i$ by gradient $\nabla\sum^m_{j=1}L_G(z_j|G_i\mathcal{N}^{(r)}_{D_i})$
            }
        }
        $\mathcal{A}^{(r)}_D\leftarrow$ Select Pareto-front discriminators under $\mathcal{F}$ based on validation set;
        
        $\mathcal{A}_D^{(r)}\leftarrow$ Crossover\&Mutation($\mathcal{A}_D^{(r)}$);
        
        $w_{G^*}\leftarrow$ the generator weights of the best GAN;
        
        $w_{G_1}=...=w_{G_P}=w_{G^*}\leftarrow$ Weight-resetting;
    }
    $D^*\sim(\beta^*,w_{D^*})\leftarrow$ the best discriminator with architecture $\beta^*$ and weights $w_{D^*}$;
    \caption{EAGAN.}
    \label{alg:eagan}
\end{algorithm}

\textbf{Weight-resetting.} The second challenge of stage-2 is that the one-to-one training strategy does not fully guarantee a fair comparison between different discriminators. Since $P$ generators are trained independently, each generator will have different weights after a round of one-to-one training, presented with different colors (see Fig. \ref{fig:EAGAN} (right)). If some generators have mode collapse due to combination with unsuitable discriminators, then subsequent discriminators paired with these generators will obtain unfair and biased estimation. To alleviate this problem, we propose the \textit{weight-resetting} strategy, which is to first copy the weights of best generator in the current round, and then initialize all generators in the next round with the copied weights. In the first round, all generators are initialized with the weights of $G^*$ found in stage-1. In summary, the one-to-one training strategy allows each discriminator to be paired with an independent generator, and the weight-resetting strategy ensures a fair comparison between different discriminators and alleviates the instability of GAN training.

% because all generators have the same architecture and are reset with the same weights at the beginning of each round.

\subsection{Architecture Evolution}\label{sec:evolution}

% Alg. \ref{alg:eagan} summarizes the detailed procedure of EAGAN algorithm. EAGAN decouples the search of G and D into two stages. Each search stage comprises $R$ rounds of evolution. In every round, a group of $P$ architectures, called \textit{population} ($\mathcal{A}=\{\mathcal{N}_1,\mathcal{N}_2,...,\mathcal{N}_P\}$), are maintained and trained. The parent individuals are used to produce a fixed number of new architectures (\textit{offspring}) via three evolution steps: selection, crossover, and mutation.

% We first introduce the encoding scheme for the search space and then describe three different evolution operations.

% As shown  in Fig. \ref{fig:EAGAN}, generators and discriminators are searched 

% \subsubsection{Encoding}

% The architectures of SuperNet-G and SuperNet-D are built by stacking multiple Up-Cells or Down-Cells. The cell structure is determined by edges, i.e., each edge will activate only one operation from multiple candidates. We encode each operation by a one-hot sequence, e.g., the bilinear interpolation is encoded as [0,1,0] in three upsampling operations. Accordingly, each architecture can be encoded as a set of one-hot sequences. In other words, searching the architecture is transformed to searching a set of one-hot sequences for different cells.

As shown in Fig. \ref{fig:EAGAN}, after weights training, stage-1 and stage-2 perform the same steps to evolve generators and discriminators, respectively. To simplify notations, we use $\mathcal{N}$, $\mathcal{N}_i$, and $\mathcal{A}$ to denote the SuperNet, the $i$-th subnet, and population, of candidate generators (stage-1) and discriminators (stage-2), respectively.

% \subsubsection{Selection.}

\textbf{Selection.} This step is equivalent to Eq. (\ref{eq:stage1_eq1}) of stage-1 and Eq. (\ref{eq:stage2_eq1}) of stage-2. In our work, we use IS \cite{IS} and FID \cite{FID} metrics to evaluate the performance of individual (i.e., subnet). FID is inversely correlated with IS, so we adopt the \textit{non-dominated sorting strategy}~\cite{nsga2} as the value function to produce the Pareto-front individuals during each round. An individual $\mathcal{N}_i$ is said to be dominated by another individual $\mathcal{N}_j$  when Eq.~(\ref{eq:domination}) satisfies.

% Specifically, an individual $\mathcal{N}_j$ is said to dominate another individual $\mathcal{N}_i$ if and only if there is no objective of $\mathcal{N}_j$ worse than that objective of $\mathcal{N}_i$ and there is at least one objective of $\mathcal{N}_j$ is better than that objective of $\mathcal{N}_i$, i.e., when Eq.~(\ref{eq:domination}) satisfies.

\begin{equation}
\begin{array}{ll}
\mathcal{F}_{k}\left(\mathcal{N}_i\right) \geq \mathcal{F}_{k}\left(\mathcal{N}_j\right) & \forall k \in\{1, \ldots, K\} \\
\mathcal{F}_{k}\left(\mathcal{N}_i\right)>\mathcal{F}_{k}\left(\mathcal{N}_j\right) & \exists k \in\{1, \ldots, K\}
\end{array}
\label{eq:domination}
\end{equation}

\noindent where $\mathcal{F}_k$ indicates the objective (e.g., FID, and $\frac{1}{IS}$\footnote{The higher the IS value, the better the GAN performance.}). We split the population with $P$ individuals into a number of disjoint subsets (or ranks) $\Omega=\{\Omega_0, \Omega_1,...\}$ by comparing the number of times each individual being dominated by other individuals, where the length of $\Omega$ and each subset may be different for each search round. After non-dominated sorting, individuals in the same subset are regarded as equally important and better than those in a larger rank. For example, the individuals in the subset $\Omega_0$ outperform all other subsets of individuals. Finally, we sequentially select $\frac{P}{2}$ individuals from lower to higher ranks.
% , forming the Pareto-front individuals of the current round.

% Likewise, the individuals in the $G_l$ list are the worst-performing ones and dominated by $l$ other individuals.

% \subsubsection{Crossover\&Mutation.}
\textbf{Crossover\&Mutation.} As detailed in Sec. \ref{sec:searchspace}, the architecture of each subnet is encoded by a set of one-hot sequences, where the one-hot sequence indicates an edge and the position of 1 indicates the candidate operation activated on that edge. Thus, the basic unit of crossover and mutation is the one-hot sequence. We set $\frac{P}{2}$ Pareto-front individuals obtained from the selection step as parents. Then, we repeatedly perform crossover and mutation on these parents with probabilities of 0.3 and 0.5, respectively, until we generate $\frac{P}{2}$ new individuals. For crossover, we randomly choose two parents and exchange a single one-hot sequence (i.e., an edge). For mutation, we also randomly choose the one-hot sequence of an edge and change the position of 1 on it.

% Fig.~\ref{fig:cross_mutation} gives an example of crossover, where two parents exchange only the operation on edge $E_{G0}$ in Fig.~\ref{fig:searchspace}.

% \begin{figure}
%      \centering
%      \includegraphics[width=\textwidth]{crossover_mutation.pdf}
%     %  \subfigure[Crossover]{\includegraphics[width=0.23\textwidth]{crossover.pdf}}
%     %  \subfigure[Mutation]{\includegraphics[width=0.23\textwidth]{mutation.pdf}}
%     \caption{Examples of crossover and mutation. For simplicity, we only show the one-hot encoding sequences of $E_{G0}$ to $E_{G2}$ in Fig.~\ref{fig:searchspace}.}
%     \label{fig:cross_mutation}
% \end{figure}

% \subsubsection{Mutation.} To enhance the diversity of the population, we perform mutation on child individuals with a probability of 0.5. Specifically, we randomly sample only one edge and change the position of 1 in its one-hot sequence. 

% Fig.~\ref{fig:cross_mutation} shows an example of mutation, where the operation on the edge $E_{G0}$ in Fig.~\ref{fig:searchspace} is mutated from the nearest neighbor interpolation ([1,0,0]) to bilinear interpolation ([0,1,0]).

% \begin{figure*}
%     \centering
%     \subfigure[Generator]{\includegraphics[width=\textwidth]{generator.pdf}}
%     \subfigure[Discriminator]{\includegraphics[width=\textwidth]{discriminator_stl10.pdf}}
%     \caption{The architectures of the searched generator and discriminator.}
%     \label{fig:generator_architecture}
%     % \vspace{-10px}
% \end{figure*}

\section{Experiments}

% In this section, we first give our used datasets and then introduce the concrete experiment settings. Lastly, results are analyzed to show improvement of EAGAN compared with existing method.

% The experiments can be divided into two stages: search and fully-train. The search stage contains two stages, i.e., stage-1 searches generator, and stage-2 searches discriminator. The search stage is only conducted on the CIFAR-10~\cite{cifar10} dataset, and then the searched architectures are fully-trained and evaluated on both CIFAR-10 and STL-10~\cite{stl10} datasets.

% \subsection{Datasets}

% We evaluate our EAGAN on the CIFAR-10~\cite{cifar10} and STL-10~\cite{stl10} datasets. CIFAR-10 consists of 50,000 training images and 10,000 test images with 32$\times$32 resolutions. STL-10 contains 100,500 images with 96$\times$96 resolutions, but we resize them to 48$\times$48, the same setting as the previous NAS-GANs.

\subsection{Implementation Settings}

% Our implementation is based on PyTorch~\cite{paszke2019pytorch} with a single NVIDIA Tesla V100 GPU and will be open-sourced. 

% Our EAGAN uses a single NVIDIA Tesla V100 GPU.

% and takes only 1.2 GPU days to finish the search for GANs on the CIFAR-10 dataset.

% Our code is based on PyTorch~\cite{paszke2019pytorch} and will be open-sourced. Our EAGAN takes only 1.2 GPU days to finish the search on a single NVIDIA Tesla V100 GPU.

\textbf{Datasets.} Following the previous NAS-GANs \cite{autoGAN,Adversarialnas,EGAN}, we search on the CIFAR-10~\cite{cifar10} and evaluate on both CIFAR-10 and STL-10~\cite{stl10} datasets. CIFAR-10 has 50,000 training images and 10,000 test images with 32$\times$32 resolutions. STL-10 has 100,500 images with 96$\times$96 resolutions, but we resize them to 48$\times$48.

\textbf{Warm-up Stage.} We set up a warm-up stage before the start of stage-1 and stage-2 to ensure a fair competition for all candidate subnets. Specifically, all candidate operations in search space are activated uniformly and trained equally. The warm-up stage has 50 epochs. After that, we randomly sample $P$ subnets to form the first round of population.

% \subsubsection{Two-stage Search.} 

\textbf{Two-stage Search.} For both stage-1 and stage-2, we use the hinge loss~\cite{SNGAN} and Adam optimizer~\cite{adam} with an initial learning rate of 0.0002. The total number of search rounds is 18, each containing 10 epochs. The noise data is sampled from the Gaussian distribution. A population of $P=32$ individuals is trained and evolved during each round. The batch sizes for generator and discriminator are 40 and 80, respectively. Besides, we adopt a low-fidelity evaluation strategy, i.e., the number of images used to calculate FID and IS is reduced to 5,000, which greatly reduces the evaluation time and keeps the performance of the searched architectures. Stage-1 and stage-2 take 0.8 and 0.4 GPU days, respectively.
% 
% When searching for a discriminator, we first select the best generator, which is confirmed to have a higher IS and a lower FID after sufficient training, from searched population generators. Then the chosen generator architecture is fixed with discriminator searching. Other implementation setting is the same as generator searching stage. 

% \subsubsection{Fully-train Stage.} 
\textbf{Fully-train Stage.} After the two-stage search, we fully train the best-performing GAN $(G^*,D^*)$ from scratch. For the CIFAR-10 dataset, the batch size and learning rate are the same as the search stage, but the total number of training epochs is 600. For the STL-10 dataset, the batch size and the learning rate are 128 and 0.0003 for the generator, and 64 and 0.0002 for the discriminator, respectively. Following the previous NAS-GAN works~\cite{Adversarialnas,autoGAN}, we generate 50,000 images to calculate IS and FID metrics.

% as the final performance.

\subsection{Results and Analysis}

% \subsubsection{Search only Generator.}

\begin{figure*}
    \centering
    \includegraphics[width=\textwidth]{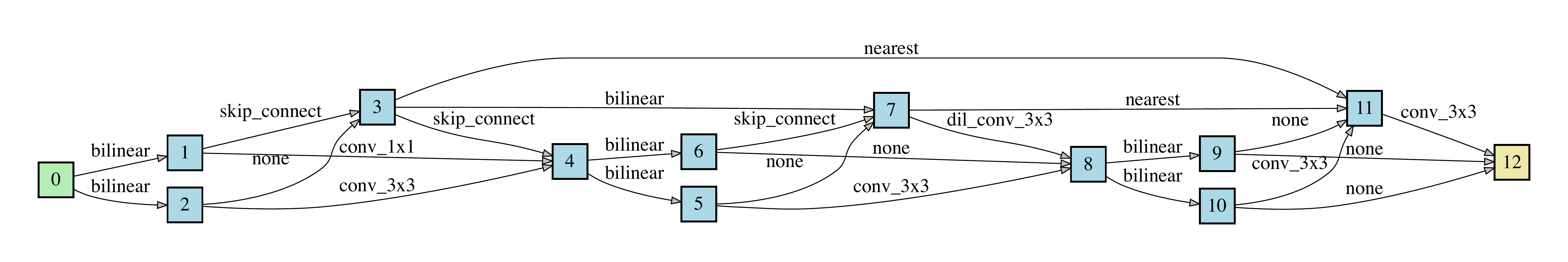}
    \caption{The architecture of the searched generator ($G_C=G_S=G^*$).}
    \label{fig:generator_architecture}
    % \vspace{-10px}
\end{figure*}

\textbf{Search only Generator (EAGAN-G).} Our searched generator $G^*$ is shown in Fig. \ref{fig:generator_architecture}. Note that the generators for the CIFAR-10 ($G_C$ with 7.14M parameters) and STL-10 ($G_S$ with 11.55M parameters) datasets have the same architecture but different input channels, so their sizes are different. We can see that 1) bi-linear operation is preferred for up-sampling, which is also observed in previous NAS-GANs~\cite{Adversarialnas,offgan}; 2) there are 6 ``None'' operations and 3 ``skip-connect'' operations among 15 total normal operations, and the normal convolution with kernel size $3\times3$ is preferred, which is probably because the low-resolution images do not need complicated convolutions to generate. The results in Table.~\ref{tab:results} show that, compared with AdversarialNAS~\cite{Adversarialnas}, our EAGAN can find a better generator with similar time overhead, given the same search space and fixed discriminator. Specifically, our discovered generator achieves a highly competitive FID (10.14) and IS (8.76$\pm$0.09) on the CIFAR-10 dataset. In terms of IS, there is a certain gap between NAS-GANs and BigGAN \cite{biggan} because BigGAN additionally introduces category information as input into the generator's architecture, while NAS-GANs only receive noise data as input. Besides, our generator $G_S$ achieves remarkable results (IS 10.02$\pm$0.11, FID=23.34) on the STL-10 dataset, showing an excellent transferability.

% In terms of IS, there is a certain gap between NAS-GANs and BigGAN \cite{biggan} because BigGAN additionally introduces category information as input into the generator's architecture, while NAS-GANs only receive noise data as input. 

% Therefore, incorporating category information and images into the search space is a worthy direction to investigate.
 
% showing an excellent transferability.

\begin{table*}[!ht]
    \centering
    \scalebox{1}{
    \begin{tabular}{c|c|c|c|c|c|c}\hline
        \multirow{2}{*}{Method} & \multirow{2}{*}{\makecell{Search\\Method}} & \multirow{2}{*}{\makecell{GPU\\Days}} & 
        \multicolumn{2}{c|}{CIFAR-10}& 
        \multicolumn{2}{c}{STL-10}  \\\cline{4-7}
        &&&IS$\uparrow$&FID$\downarrow$&IS$\uparrow$&FID$\downarrow$\\\hline
        \makecell{DCGANs~\cite{dcgan}} & \multirow{7}{*}{Manual}& \multirow{7}{*}{--} &6.64$\pm$0.14& 37.7&-- &-- \\
        \makecell{WGAN-GP~\cite{wgan-gp}} &  & &7.86$\pm$0.07& 29.3& --&-- \\
        \makecell{Progressive GAN~\cite{progressiveGAN}}  & & &  8.80$\pm$0.05& 18.33 &-- &-- \\
        \makecell{SN-GAN~\cite{SNGAN}} &  & &8.22$\pm$0.05& 21.7 &9.16$\pm$0.12 &40.1 \\
        ProbGAN~\cite{he2018probgan}&&&7.75&24.60&8.87$\pm$0.09&46.74\\
        \makecell{Improv MMD GAN\cite{improving_mmd_gan}}&&&8.29&16.21&9.34&37.63\\
        % StyleGAN-V2~\cite{stylegan2}&&&9.18&11.07&-&-\\
        BigGAN~\cite{biggan}&&&\textbf{9.22}&14.73 &-&-\\\hline
        \makecell{AGAN~\cite{agan}} & \multirow{4}{*}{RL} &1200  &8.29$\pm$0.09 & 30.5&9.23$\pm$0.08 &52.7 \\
        \makecell{AutoGAN~\cite{autoGAN}}&  &2 & 8.55$\pm$0.10&  12.42& 9.16$\pm$0.12&31.01 \\
        \makecell{E2GAN~\cite{offgan}}& &0.3 &8.51$\pm$0.13 & 11.26& 9.51$\pm$0.09&25.35 \\\hline
        \makecell{DEGAS~\cite{DEGAS}} &\multirow{5}{*}{Gradient} & 1.167 &8.37$\pm$0.08 & 12.01&9.71$\pm$0.11 &28.76 \\
        AlphaGAN~\cite{AlphaGAN} & &0.13 & 8.98$\pm$0.09&10.35&10.12$\pm$0.13&22.43\\
        AlphaGAN~\cite{AlphaGAN}$\dagger$ & & - & 8.70$\pm$0.11&15.56&-&-\\
        \makecell{AdversarialNAS~\cite{Adversarialnas}} & & 1& 7.86$\pm$0.08 & 24.04& 8.52$\pm$0.05&38.85 \\
        \makecell{AdversarialNAS~\cite{Adversarialnas}$\dagger$} & & 1 &8.74$\pm$0.07&10.87&9.63$\pm$0.19&26.98\\\hline
        DGGAN~\cite{dggan}&Heuristic&580&8.64$\pm$0.06&12.10&-&-\\\hline
        EGAN~\cite{EGAN} &\multirow{2}{*}{EA}&1.25&6.9$\pm$0.09&-&-&-\\
        EAS-GAN~\cite{EAS-GAN} & & 1 & 7.45$\pm$0.08 & 33.2 & - & 38.84 \\\hline\hline
        \textbf{EAGAN-G} &\multirow{4}{*}{EA} & 0.8 &  8.76$\pm$0.09 & 10.14 & 10.02$\pm$0.11 &  23.34\\
        \textbf{EAGAN-GD1}$\dagger$ & &0.8+0.4 & 8.81$\pm$0.10& \textbf{9.91}& \textbf{10.44$\pm$0.08}&\textbf{22.18}\\
        \textbf{EAGAN-GD2}$\dagger$ & &0.75+0.37 & 8.63$\pm$0.09& 12.84& 9.76$\pm$0.06 & 26.52 \\
        \textbf{EAGAN-GD3}$\dagger$ & &1.55+0.73 & 8.69$\pm$0.10& 10.53& 10.14$\pm$0.11&24.22\\
        \hline
    \end{tabular}
    }
    \caption{Results on the CIFAR-10 and STL-10 datasets. $\dagger$ indicates searching both generators (G) and discriminators (D).}
    \label{tab:results}
\end{table*}

% G size=11.55 MB
% D-cifar10:  D-stl10: 1.58 MB

\begin{figure*}[!h]
    \centering
    % \vspace{-30pt}
    \subfigure{\includegraphics[width=\textwidth]{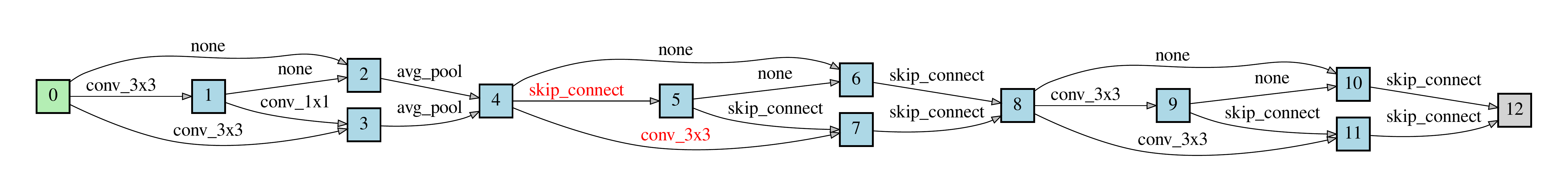}}
    \subfigure{\includegraphics[width=\textwidth]{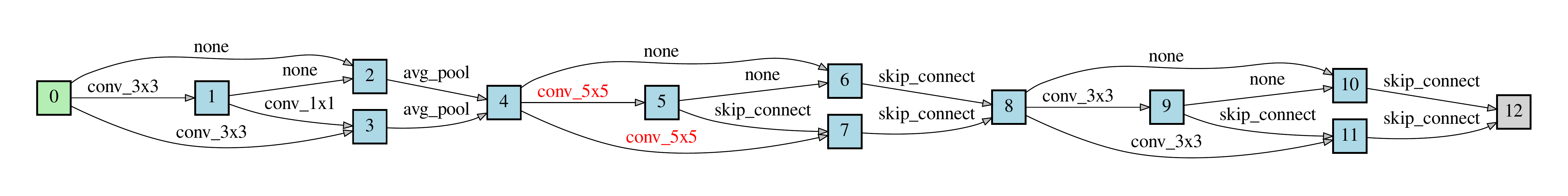}}
    \caption{The searched discriminators on CIFAR-10 (top) and STL-10 (bottom).}
    \label{fig:searched_d}
\end{figure*}

\textbf{Search both Generator and Discriminator (EAGAN-GD1).} In stage-2, we use the best generator $G^*$ found in stage-1 to help search a set of Pareto-front discriminators, from which we select the optimal discriminators for the CIFAR-10 ($D_C$ with 0.91M parameters) and STL-10 ($D_S$ with 1.58M parameters) datasets, respectively, shown in Fig.~\ref{fig:searched_d}. We can see a subtle difference (marked in red) between them, i.e., $D_S$ prefers convolutions with a larger kernel size ($5\times5$), while $D_C$ selects skip-connection and a smaller convolution. A possible reason is that the resolution of STL-10 (48$\times$48) is larger than CIFAR-10 (32$\times$32), so it needs a larger kernel size to obtain larger receptive fields.

% \begin{figure*}[!h]
%     \centering
%     % \vspace{-30pt}
%     \subfigure[The architecture of discriminator for CIFAR-10 ($D_C$).]{\includegraphics[width=\textwidth]{cifar10_D.pdf}}
%     \subfigure[The architecture of discriminator for STL-10 ($D_S$).]{\includegraphics[width=\textwidth]{discriminator_stl10.pdf}}
%     \caption{The architectures of the searched discriminators on CIFAR-10 (top) and STL-10 (bottom).}
%     \label{fig:searched_d}
% \end{figure*}

% We obtain two GANs (i.e., ($G_C,D_C$) and ($G_C,D_S$)) and retrain them from scratch. From Table.~\ref{tab:results}, w

\begin{figure*}[h!]
    \centering
    \subfigure[CIFAR-10]{\includegraphics[width=0.49\textwidth]{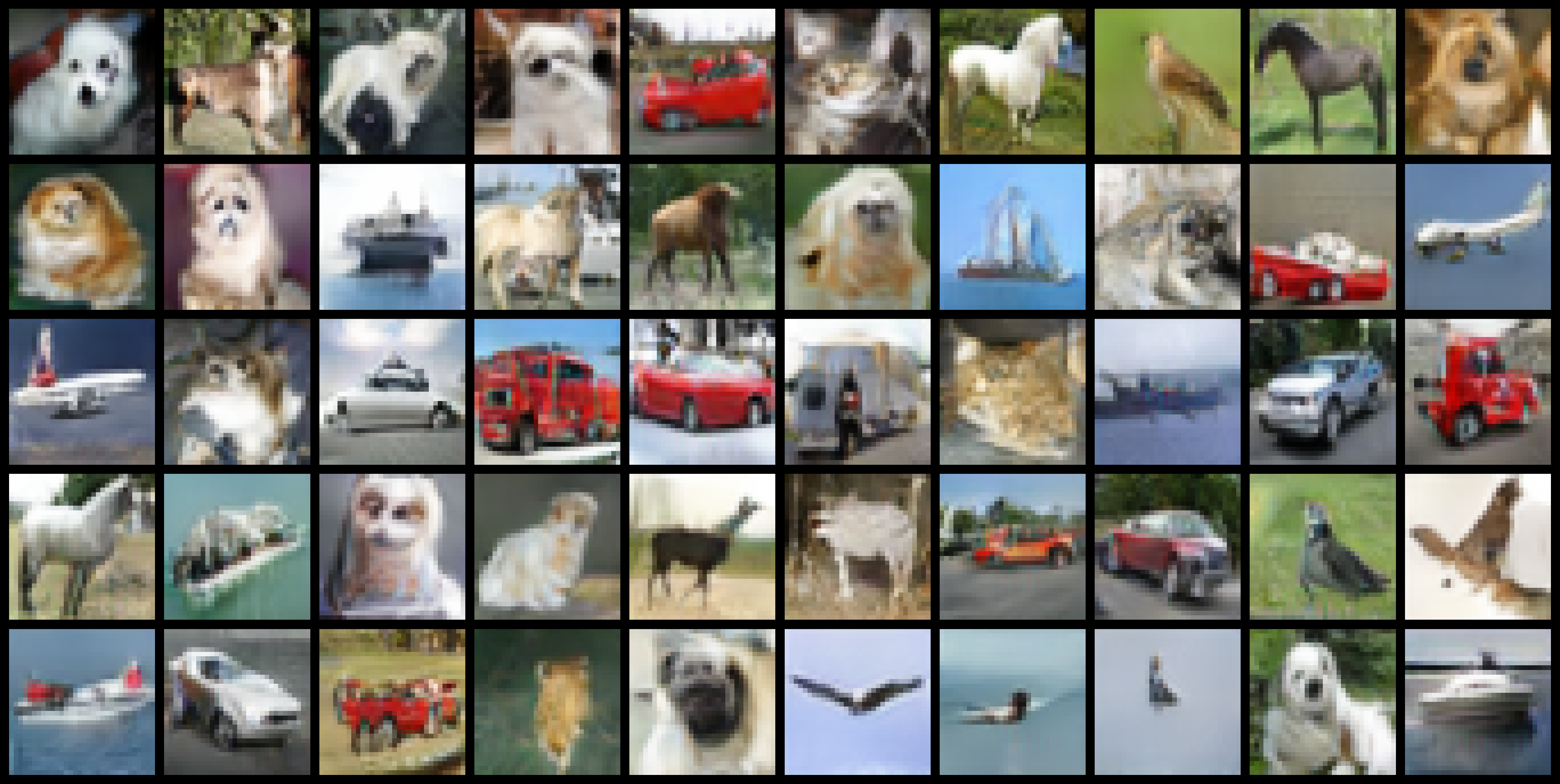}}
    \subfigure[STL-10]{\includegraphics[width=0.49\textwidth]{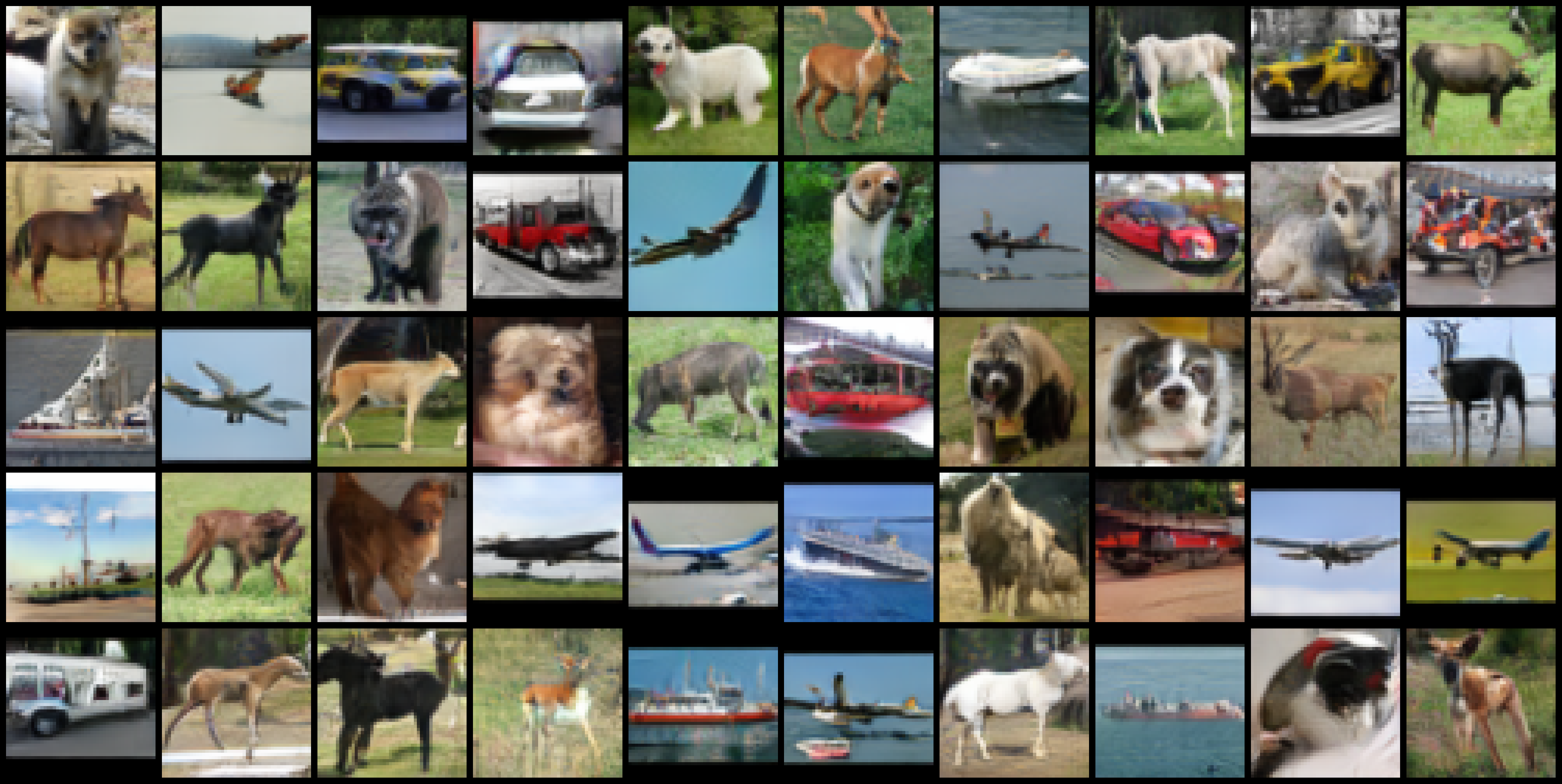}}
    \caption{The generated images by EAGAN in random without cherry-picking.}
    \label{fig:generated_imgs}
\end{figure*}

After two-stage search, we retrain two GANs (i.e., ($G_C,D_C$) and ($G_C,D_S$)) on the CIFAR-10 and STL-10 datasets, respectively, and report their results in Table.~\ref{tab:results}. We can see that none of existing NAS-GANs can guarantee to find excellent GANs in both search scenarios: (a) searching only generators; and (b) searching both generators and discriminators. For example, AdverearialNAS~\cite{Adversarialnas} performs poorly (IS=7.86$\pm$0.08, FID=24.04) in scenario (a), and AlphaGAN~\cite{AlphaGAN} suffers from instability in scenario (b), as its performance drops significantly from (IS=8.89$\pm$0.09, FID=10.35) in scenario (a) to (IS=8.70$\pm$0.11, FID=15.56) in scenario (b). However, our EAGAN performs well in both search scenarios, and the discriminators searched in stage-2 can further improve the performance of the optimal generator discovered in stage-1. Specifically, we achieve a competitive IS value (8.81$\pm$0.10) and the best FID (9.91) on the CIFAR-10 dataset. Besides, our EAGAN achieves remarkable performance (IS=10.44$\pm$0.08, FID=22.18) on the STL-10 dataset, which outperforms the existing NAS-searched GANs. In Fig.~\ref{fig:generated_imgs}, we present 50 images randomly generated by generators trained on the CIFAR-10 and the STL-10 datasets without cherry-picking, respectively. The generated images are of rich diversity and high quality.

\subsection{Ablation Study}\label{sec:ablation}
% \subsubsection{Ablation Study}

% To further evaluate the effectiveness of the proposed EAGAN, we conduct a series of ablation study on CIFAR-10. First, 
% \subsubsection{Two-stage Searching.}

\textbf{Search G or D first?} EAGAN searches G first and then searches D. \textit{What about search D first?} Our experiments show that searching D first in stage-1 will make the searched D much stronger than candidate G in stage-2, which in turn causes the gradients of G to vanish. Thus, we should search G first.

\textbf{Initialize different D in stage-1.} Our above experiment (i.e., EAGAN-GD1) uses the discriminator of \cite{Adversarialnas} in stage-1. We further implement two experiments to explore the effect of initializing different D in stage-1. EAGAN-GD2 uses a simple network with 0.92M parameters, comprising five normal convolutions and a linear layer, as the initial D in stage-1. EAGAN-GD3 is to repeat the two-stage search several times, i.e., the optimal D of the previous stage-2 is set as the initial D of the next stage-1. From Table. \ref{tab:results}, we can see that both EAGAN-GD2 and EAGAN-GD3 achieve competitive results on the CIFAR-10 and STL-10 datasets, indicating that EAGAN does not require strong prior knowledge to design the initial state of D and that searching once is sufficient to find good models, balancing search overhead and model performance.

% Although the final EAGAN-GD3 perform worse than above two, its acceptable result can also confirm that our paradigm does not necessary require strong prior knowledge to initialize discriminator and is robust.

% 
% In above experiments, we search an optimal $G^*$ in stage-1 and then search an optimal $D^*$ in stage-2. We further repeat a two-stage search, where $D^*$ is set as the initial discriminator in stage-1. The performance of the final searched GAN is IS

\begin{figure}
    \centering
    \includegraphics[width=0.7\textwidth]{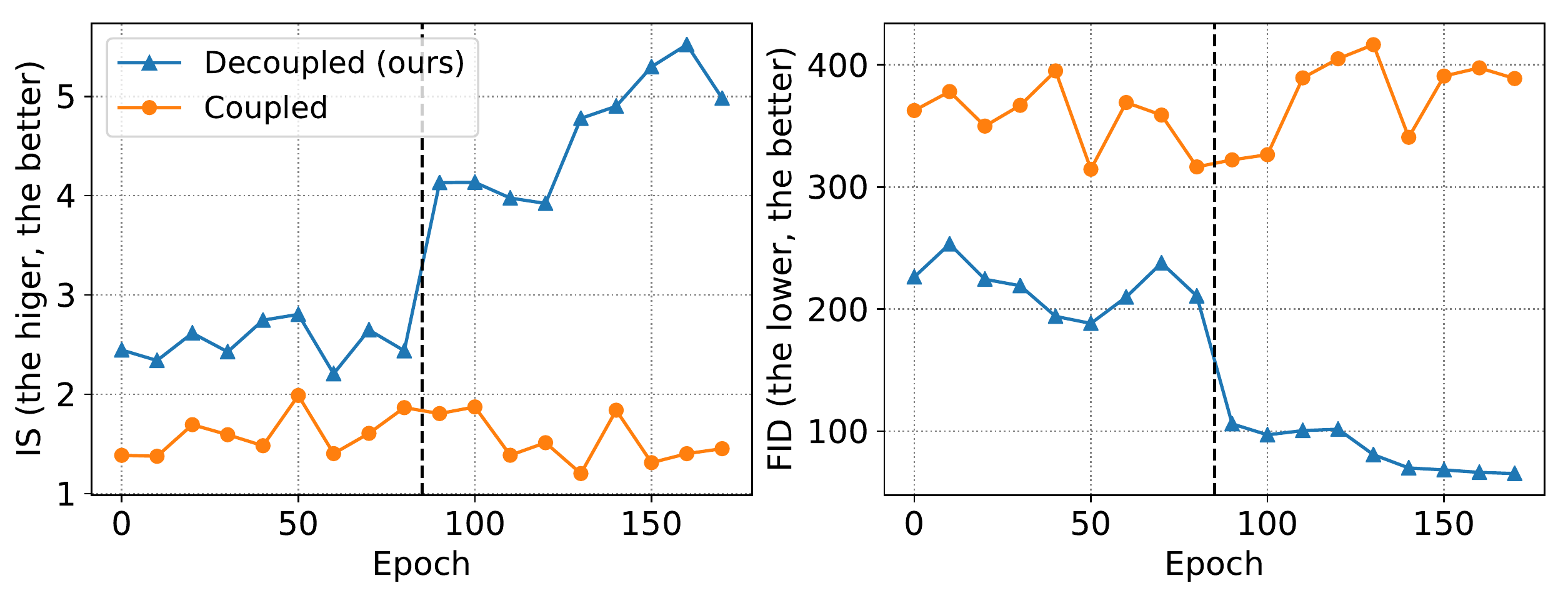}
    % \subfigure[IS in search]{\includegraphics[width=0.225\textwidth]{cifar10_is_compair.pdf}}
    % \subfigure[FID in search]{\includegraphics[width=0.225\textwidth]{cifar10_fid_compair.pdf}}
    
    \caption{Learning curves when generators and discriminators are coupled/decoupled. The dashed line indicates the boundary between the two decoupled stages of EAGAN. }
    \label{fig:converge_compair}
\end{figure}

\textbf{Decoupled vs. Coupled.} To validate the effectiveness of our decoupled search method, we perform a coupled search experiment as the baseline, i.e., the architectures of G and D are evolved simultaneously for each search round. Fig.~\ref{fig:converge_compair} presents the learning curves of the baseline and our EAGAN, which shows that coupled search is unstable as it fluctuates throughout the search. In contrast, the overall performance of our decoupled search is better and significantly improved, especially in stage-2 of searching discriminators. Besides, the decoupled search also fluctuates in stage-1 due to the competition among candidate generators incurred by the weight-sharing strategy, and how to address the negative impact of weight-sharing is still an open problem \cite{xie2021weightsharing}.

% may be caused by different candidate generators sharing weights and competing with each other in optimizing the weights; and that the overall performance of EAGAN is obviously better than the baseline. In terms of stage-2 of searching discriminators, the performance of EAGAN improves significantly, while the baseline is still unstable.

\begin{figure}
    \centering
    \includegraphics[width=0.7\textwidth]{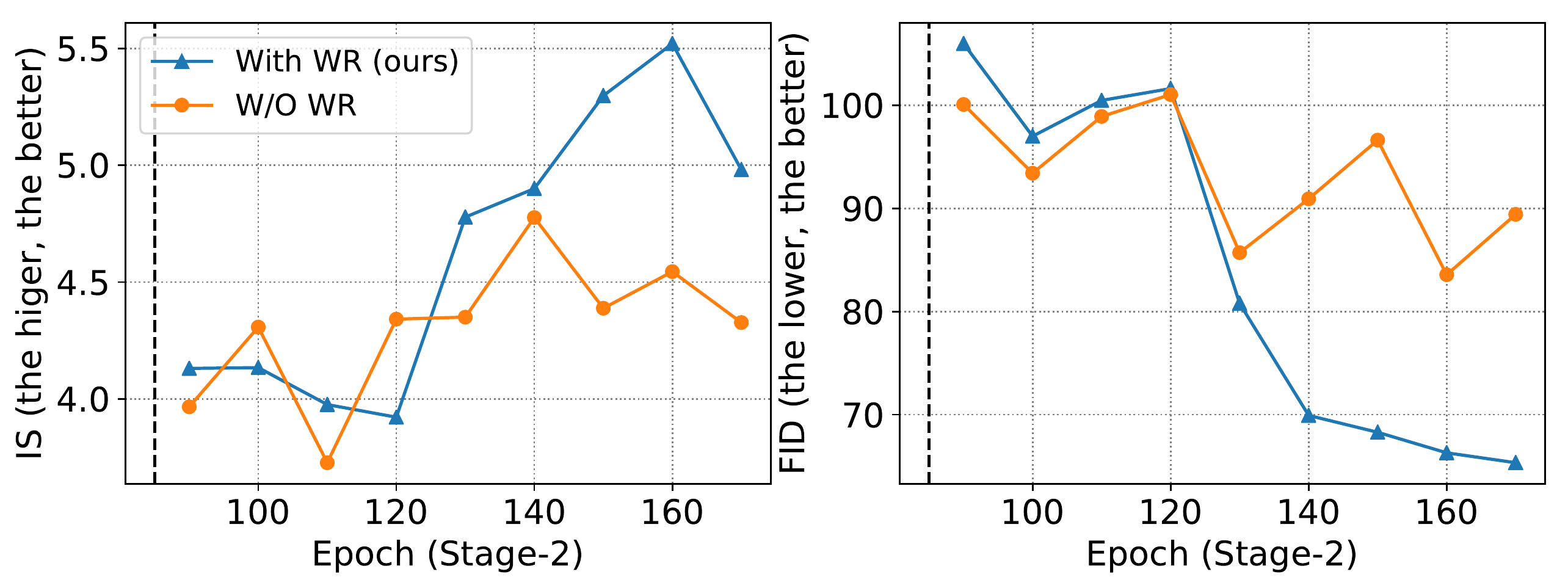}
    % \subfigure[IS in search]{\includegraphics[width=0.225\textwidth]{cifar10_is_compair.pdf}}
    % \subfigure[FID in search]{\includegraphics[width=0.225\textwidth]{cifar10_fid_compair.pdf}}
    
    \caption{Learning curves with and without (W/O) the weight-resetting (WR) strategy in stage-2.}
    \label{fig:effect_wr}
\end{figure}

% \subsubsection{Weight-resetting Strategy.}

\textbf{Weight-resetting Strategy.} We conduct another experiment on the CIFAR-10 dataset, which differs from our EAGAN only in that the weights of $P$ generators in stage-2 are continuously and independently trained without weight-resetting (WR) strategy. Fig.~\ref{fig:effect_wr} presents the learning curves with and without the WR strategy in stage-2, which shows that our proposed WR strategy can effectively enhance the stability of GAN training and obtain better IS and FID scores in stage-2 of searching discriminators.

% \begin{figure}
%     \centering
%     \includegraphics[width=0.49\textwidth]{cifar10_is_fid_compair.pdf}
%     % \subfigure[IS in search]{\includegraphics[width=0.225\textwidth]{cifar10_is_compair.pdf}}
%     % \subfigure[FID in search]{\includegraphics[width=0.225\textwidth]{cifar10_fid_compair.pdf}}
    
%     \caption{The curve of IS and FID during search.}
%     \label{fig:converge_compair}
% \end{figure}

\section{Conclusion \&  Future Work}

This paper proposes an efficient two-stage evolutionary algorithm-based NAS framework to search GANs, namely EAGAN.  We demonstrate that decoupling the search of the generator and discriminator into two stages can significantly improve the stability of searching GANs via the GAN training strategies (many-to-one and one-to-one) tailored for both stages and the weight-resetting strategy. EAGAN is very efficient and takes 1.2 GPU days to finish the search on CIFAR-10. Our searched GANs achieve competitive performance (IS and FID) on the CIFAR-10 dataset and outperform previous NAS-GANs on the STL-10 dataset. 

We believe our work deserves more in-depth study and may benefit other potential fields. For example, our decoupled paradigm and tailored training strategies are well suited for large-scale parallel search when architectures require adversarial training. Further, we shall investigate reducing the interference of weight-sharing in search and explore high-resolution generative tasks.

\noindent\textbf{Acknowledgements.} Thanks to the NVIDIA AI Technology Center (NVAITC) for providing the GPU cluster to support our work. BH was supported by the NSFC Young Scientists Fund No. 62006202, Guangdong Basic and Applied Basic Research Foundation No. 2022A1515011652, RGC Early Career Scheme No. 22200720, RGC Research Matching Grant Scheme No. RMGS2022\_11\_02 and HKBU CSD Departmental Incentive Grant.

% In the future, we shall investigate how to reduce the interference of weight sharing on GAN search and study incorporating category and image information into the search space.

% In the future, we will make further improvements in the following aspects: incorporating category and image information into the search space, exploring high-resolution generative tasks, and finding better and cheaper metrics.

% We demonstrate that decoupling the search of generator and discriminator into two stages can greatly improve the stability of searching GANs. Specifically, we propose many-to-one and one-to-one training strategies tailored for stage-1 and stage-2, respectively. The proposed weight-resetting strategy can effectively alleviate the negative impact of some generators with mode collapse when searching discriminators, and enable a fair and unbiased evaluation for different discriminators. EAGAN is very efficient and takes only 1.2 GPU days to finish the search on CIFAR-10. Our searched GANs have advanced transferability and achieve remarkable performance (IS and FID) on STL-10 datasets. In the future, we will make further improvements in the following aspects: incorporating category information into the search space, exploring high-resolution generative tasks, and finding better and cheaper metrics.

\bibliographystyle{splncs04}
\bibliography{eccv}

% \clearpage\mbox{}Page \thepage\ of the manuscript.
% \clearpage\mbox{}Page \thepage\ of the manuscript.

% This is the last page of the manuscript.
% Now we have reached the maximum size of the ECCV 2022 submission (excluding references).
% References should start immediately after the main text, but can continue on p.15 if needed.

% \clearpage
% \bibliography{egbib}
\end{document}